\documentclass[conference]{IEEEtran}
\usepackage{times}

\usepackage[numbers]{natbib}
\usepackage{multicol}
\usepackage[bookmarks=true]{hyperref}
\usepackage{graphicx}
\usepackage{float}
\usepackage{stfloats}
\usepackage{amsmath}
\usepackage{amsfonts}
\usepackage{xcolor}
\usepackage{soul}
\sethlcolor{yellow}
\usepackage[skip=4pt]{caption}

\usepackage[T1]{fontenc}
\usepackage[utf8]{inputenc}

\let\labelindent\relax
\usepackage{enumitem}

\usepackage{threeparttable}
\usepackage{booktabs}
\usepackage{tabularx}
\usepackage{siunitx}
\usepackage{multirow}
\usepackage{array}

\usepackage{booktabs}
\usepackage{tabularx}
\usepackage{threeparttable}
\usepackage{siunitx}
\usepackage{xcolor}
\usepackage{booktabs}
\usepackage{tabularx}

\begin{document}

\title{\scalebox{0.95}[1]{\begin{tabular}{@{}c@{}}AeroMap3D: Anchoring Monocular UAV 6-DoF\\Localization to Visual-Geometric-Semantic Map Priors\end{tabular}}}

\author{
\IEEEauthorblockN{Zhiyun Deng\textsuperscript{1,2},
Luis Sentis\textsuperscript{1,2}}
\IEEEauthorblockA{\textsuperscript{1}The University of Texas at Austin
\quad \textsuperscript{2}AIVE AI Systems}
}

\maketitle

\begin{abstract}
We present AeroMap3D, a monocular 6-DoF UAV localization system that anchors onboard imagery to visual, geometric, and semantic map priors for GNSS-denied navigation. AeroMap3D addresses two fundamental challenges in map-referenced aerial localization: the cross-view discrepancy between UAV imagery and satellite maps, and the structural inconsistency between bare-earth digital elevation models (DEMs) and urban scenes. First, we introduce a lightweight adapter that enables a dense matcher pretrained on internet-scale generic data to perform reliable UAV-to-map registration without finetuning. By estimating the scale ratio and yaw offset between the UAV image and map tile, the adapter removes the dominant geometric misalignment induced by altitude, camera field of view, and heading before dense correspondence estimation. Second, AeroMap3D lifts 2D UAV--map correspondences onto DEM terrain while using OpenStreetMap annotations to reject semantically unreliable matches before RANSAC--PnP pose estimation, thereby reducing errors caused by unmodeled building heights and off-nadir structures. Delayed map-based pose measurements are further fused with relative-motion priors using a delayed-state EKF for continuous trajectory estimation. Without UAV-Terra3D retraining or tuning, AeroMap3D localizes all trajectories across eight Austin sites within 50~m and achieves 5.88~m mean 3D error over 55~km of flight.
\end{abstract}

\IEEEpeerreviewmaketitle

\section{Introduction}
\label{sec:introduction}
GNSS-denied UAV navigation needs a persistent world-frame reference to keep long trajectories metrically anchored. Visual odometry provides relative motion but accumulates drift \cite{nister2004visual}, and SLAM reduces drift mainly through loop closures in a locally built map~\cite{mur2017orb}. Geo-referenced imagery and elevation data offer a complementary external anchor without requiring a return to previously observed places.

\begin{figure}[t]
    \centering
    \includegraphics[width=1\linewidth]{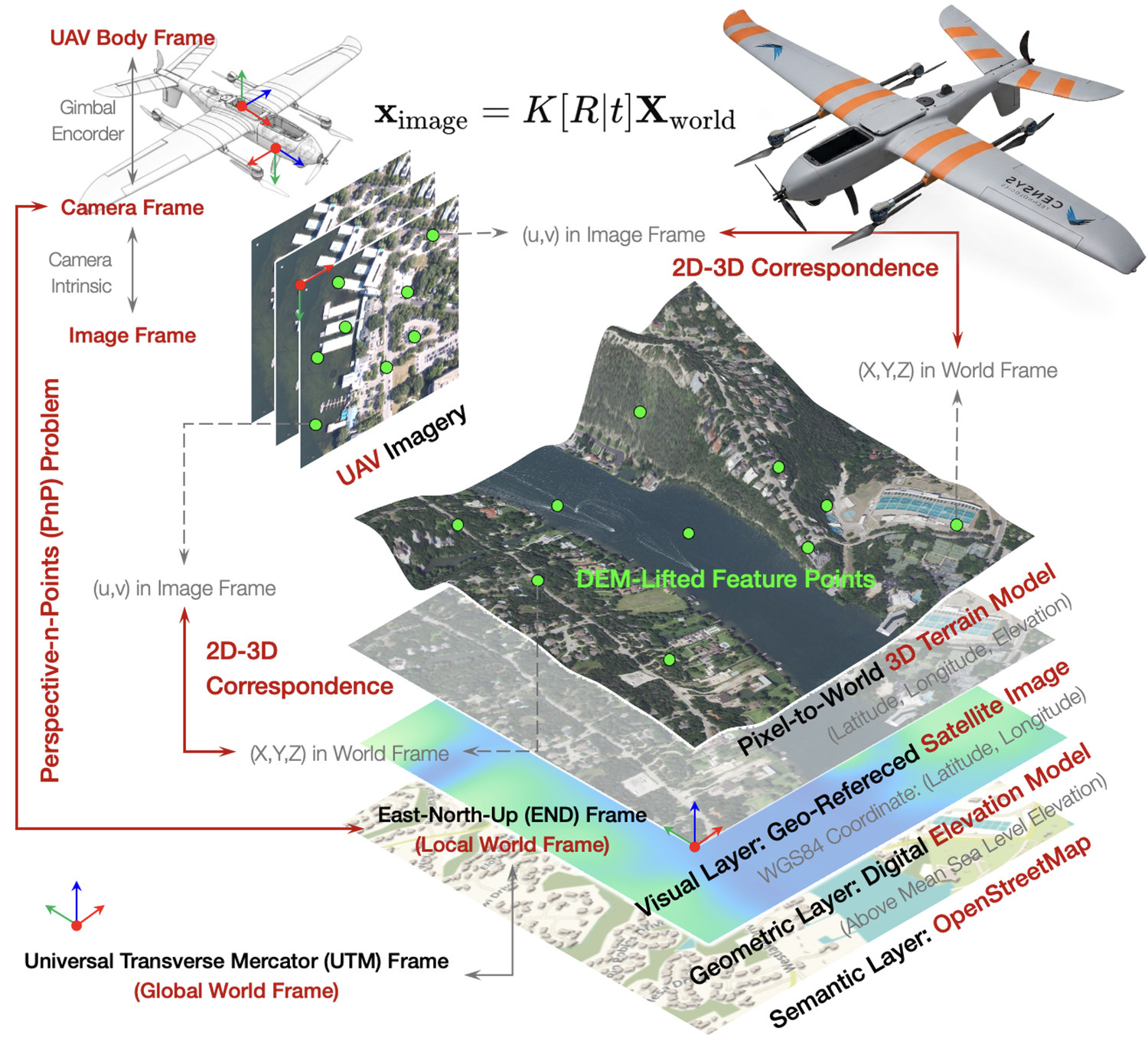}
    \caption{\textbf{Overview of AeroMap3D.} Compared with prior 6-DoF UAV localization systems that rely on textured 3D maps, DSMs, or render databases, AeroMap3D uses widely available satellite imagery, bare-earth DEM, and OSM layers, enabling better generalization to unexplored areas. A scale/yaw adapter and OSM masking condition UAV--map correspondences before DEM lifting and standard RANSAC--PnP.}
    \label{fig:overview}
    \vspace{-2em}
\end{figure}

The key obstacle is the map representation. Retrieval-based localization scales to large areas by matching UAV images to geo-tagged satellite tiles~\cite{arandjelovic2016netvlad,li2025geovins,lindenberger2026scaling}, but its accuracy is bounded by reference-database sampling. Metric 6-DoF pose estimation instead uses Perspective-$n$-Point (PnP) over 2D--3D correspondences~\cite{opencv_solvepnp_forum}, requiring map geometry that supplies elevation or depth. Existing systems obtain this geometry from pose-sampled renderings, textured 3D reconstructions, or orthophoto--digital surface model (DSM) pairs~\cite{chen2021real,wu2024uavd4l,cheng2026pilot,dhaouadi2025ortholoc}. These representations can be accurate, but they often require site surveys, region-specific data acquisition, and substantial offline construction, limiting rapid deployment in previously unseen regions. A practical alternative is therefore to localize against widely available public maps.

This paper asks whether accurate UAV localization can instead be built from public geospatial layers: geo-referenced satellite imagery \cite{usda_naip}, bare-earth DEMs \cite{usgs_3DEP}, and OpenStreetMap (OSM)~\cite{OpenStreetMap}. This reference avoids site-specific 3D reconstruction but creates two coupled correspondence failures. First, UAV and north-up satellite views differ in scale and yaw because of altitude, field of view, and heading; these shifts remain difficult for modern matchers~\cite{sun2021loftr,edstedt2024roma,edstedt2025roma}, and visual yaw estimation avoids reliance on magnetometers that may be biased~\cite{opromolla2020magnetometer,stewart2015magnetometer}. Second, a bare-earth DEM assigns terrain height to every map pixel. A correct image match on a roof or facade therefore becomes a wrong 2D--3D correspondence, and neighboring structure matches can share a coherent height bias that survives RANSAC \cite{chum2008optimal} rather than appearing as independent outliers.

We present \textbf{AeroMap3D} (Fig.~\ref{fig:overview}), a monocular UAV localization framework that derives metric 6-DoF pose estimates from public visual, geometric, and semantic map priors. It targets low-altitude urban flight, where buildings occupy much of the camera view and their omission from bare-earth DEMs most affects pose estimation. Given a coarse map initialization, a lightweight Siamese adapter estimates UAV-to-map scale and yaw, rectifying the pair before a frozen dense matcher. AeroMap3D then rejects matches in OSM-annotated building regions, lifts the remaining UAV--map matches into terrain-supported 2D--3D correspondences, recovers camera pose with standard RANSAC--PnP~\cite{chum2008optimal,opencv_solvepnp_forum}, and fuses map-anchored pose measurements with relative-motion priors in an EKF for continuous localization.

\begin{table*}[t]
\centering
\begin{threeparttable}
\scriptsize
\sffamily
\renewcommand{\arraystretch}{1.3}
\begin{tabular*}{\textwidth}{@{\extracolsep{\fill}} l c c c c c c c c l @{}}
\toprule
\multirow{2}{*}{\textbf{Dataset}}
& \multicolumn{4}{c}{\textbf{UAV Video}}
& \multicolumn{4}{c}{\textbf{Reference Map}}
& \multirow{2}{*}{\textbf{Focus}} \\
\cmidrule(lr){2-5}
\cmidrule(lr){6-9}
& \textbf{\# Traj.}
& \textbf{Camera Angle}
& \textbf{Flight Altitude}
& \textbf{Intrinsics}
& \textbf{Satellite}
& \textbf{Elevation}
& \textbf{Semantic}
& \textbf{3D Coverage}
& \\
\midrule

ALTO\tnote{1}~\cite{cisneros2022alto}
& 2
& Top-Down
& Almost Const.
& --
& Satellite
& --
& --
& --
& Image Retrieval \\

UAV-VisLoc\tnote{1}~\cite{xu2024uav}
& 11
& Top-Down
& Almost Const.
& --
& Satellite
& --
& --
& --
& Image Retrieval \\

AerialVL~\citep{he2024aerialvl}
& 11 (70\,km)
& Top-Down
& 120\,m / 200\,m
& --
& Satellite
& --
& --
& --
& Image Retrieval \\

UAVD4L~\citep{wu2024uavd4l}
& 5 (10\,km)
& Various
& Various
& Yes
& Satellite
& 3D Model
& --
& 2.5\,km$^2$
& Metric Localization\tnote{2} \\

OrthoLoC\tnote{1}~\citep{dhaouadi2025ortholoc}
& 52
& Top-Down
& Various
& Yes
& Satellite
& DSM\tnote{3}
& --
& 8.9\,km$^2$
& Metric Localization\tnote{2} \\

\textbf{UAV-Terra3D}
& \textbf{20 (55\,km)}
& \textbf{Top-Down}
& \textbf{Various}
& \textbf{Yes}
& \textbf{Satellite}
& \textbf{DEM}\tnote{3}
& \textbf{OSM}
& \textbf{22.4\,km$^2$}
& \textbf{Metric Localization}\tnote{2} \\

\bottomrule
\end{tabular*}
\begin{tablenotes}
\item[1] These datasets provide isolated images or discontinuous
trajectory samples rather than continuous UAV video sequences.
\item[2] Metric localization estimates a precise, geo-referenced spatial coordinate (x, y, z) and orientation (roll, pitch, yaw) within a given map using real-world units.
\item[3] Digital Elevation Models (DEMs) generally have broader availability than Digital Surface Models (DSMs). For example, the USGS 3D Elevation Program (3DEP) \cite{usgs_3DEP} provides high-resolution DEM products across the United States, whereas its standard IfSAR-derived DSM product is primarily available for Alaska.
\end{tablenotes}
\caption{\textbf{Comparison of representative UAV datasets.}
UAV-Terra3D provides 55\,km of continuous real-world trajectories with co-registered visual-geometric-semantic map priors across 22.4\,km$^2$, offering broader 3D coverage than prior datasets.}
\label{tab:dataset-comparison}
\end{threeparttable}
\vspace{-1em}
\end{table*}

\begin{table}[t]
\centering
\scriptsize
\sffamily
\setlength{\tabcolsep}{3pt}
\renewcommand{\arraystretch}{1.12}
\begin{tabularx}{\columnwidth}{@{} l
    >{\raggedright\arraybackslash}X
    >{\raggedright\arraybackslash}X @{} }
\toprule
\textbf{Method} & \textbf{Required Map} & \textbf{Deployment Requirement} \\
\midrule
Chen et al.~\cite{chen2021real}
& Satellite image + topography
& Offline pose-sampled RGB--D and descriptor database \\

PiLoT~\cite{cheng2026pilot}
& High-fidelity geo-referenced 3D map
& A textured surface model must already exist \\

OrthoTrack~\cite{dhaouadi2026orthotrack}
& Orthophoto + DSM
& A DSM must be available for the deployment region \\

\textbf{AeroMap3D}
& \textbf{Satellite image + bare-earth DEM + OSM}
& \textbf{No textured 3D model, DSM, or pose-sampled database} \\
\bottomrule
\end{tabularx}
\caption{\textbf{Map requirements for UAV 6-DoF localization.}
Prior methods depend on region-specific pose-sampled rendering databases, textured 3D models, or DSMs. AeroMap3D instead constructs localization maps remotely using public satellite imagery, bare-earth DEMs, and OSM data, without a textured model or pose-sampled database.}
\label{tab:method-comparison}
\vspace{-2em}
\end{table}

Our main contributions are:

\begin{enumerate}[label=\arabic*),nosep,leftmargin=*]

\item
We formulate public satellite imagery, bare-earth DEMs, and OSM semantics as complementary priors for monocular 6-DoF UAV localization. We reveal a systematic failure of conventional DEM lifting: visually correct correspondences on elevated structures produce geometrically biased constraints when assigned terrain elevation. Accounting for this visual--geometric inconsistency increases single-frame localization success from 88.24\% to 95.69\% over a geometry-only RANSAC--PnP baseline.

\item
We introduce a lightweight scale--yaw adapter that extends the generalization capability of RoMav2 to UAV-to-map registration. Rather than retraining the dense matcher, the adapter isolates the dominant cross-view variation induced by altitude, camera field of view, and heading. Trained entirely from synthetic satellite-derived pairs without real UAV imagery, manual labels, or site-specific tuning, it raises the registration success rate from 62.4\% to 99.2\% with only 35~ms of additional edge-device latency.

\item
We demonstrate AeroMap3D as a complete map-anchored localization system for drift-tolerant UAV navigation through 55~km of real-world flight over 22.4~km$^2$ of mapped area. We further release UAV-Terra3D, a public benchmark containing calibrated UAV imagery, reference trajectories, and spatially aligned satellite, DEM, and OSM layers for evaluating map-anchored UAV localization.

\end{enumerate}

\section{Related Work}
\label{sec:related-work}

\subsection{Map-Referenced Place Recognition via Image Retrieval}
\label{subsec:retrieval-related-work}

Image retrieval localizes a UAV by matching a query image to geo-tagged satellite or orthophoto map tiles. NetVLAD established learned global descriptors for place recognition~\cite{arandjelovic2016netvlad}, while recent methods use vision foundation models and specialized aerial descriptors to improve robustness to viewpoint, scale, and appearance changes~\cite{yang2025dinov2,li2025geovins,lu2025selavpr++,tzachor2025effovpr,zhang2025efficient}.
Chen et al.~\cite{chen2021real} recover 6-DoF pose by retrieving a pose-sampled RGB--D rendering and refining it through 2D--3D feature correspondences. However, this approach requires an offline rendering database for each anticipated flight region and was demonstrated only within candidate-pose areas of $400 \times 400$~m, limiting its spatial scalability.
GeoVINS uses retrieval as a geographic correction in visual--inertial estimation~\cite{li2025geovins}, and large-scale retrieval pipelines have reached continent-level coverage~\cite{lindenberger2026scaling}. These methods scale well, but their localization accuracy is ultimately tied to database sampling density.

\subsection{Map-Referenced UAV 6-DoF Localization via PnP}
\label{subsec:pnp-related-work}

Structure-based localization estimates camera pose from 2D--3D correspondences with PnP~\cite{opencv_solvepnp_forum} and RANSAC~\cite{chum2008optimal}. Modern matchers such as SuperGlue~\cite{sarlin2020superglue}, LoFTR~\cite{sun2021loftr}, and RoMa~\cite{edstedt2024roma,edstedt2025roma} improve the 2D matching stage under large appearance changes.
To recover metric 6-DoF pose, these systems must associate image pixels with 3D map points; they differ primarily in how this geometry is supplied. UAVD4L pre-renders synthetic RGB--D views from a site-specific textured 3D reconstruction~\cite{wu2024uavd4l}. PiLoT instead registers live imagery against RGB--D views rendered online from a high-fidelity 3D map~\cite{cheng2026pilot}. OrthoTrack lift orthophoto correspondences through co-registered DSMs for single-frame and continuous trajectory estimation, respectively~\cite{dhaouadi2026orthotrack}. In each case, successful lifting assumes that the reference height at a matched map pixel corresponds to the surface visible in the UAV image.
AeroMap3D instead uses bare-earth DEMs available through large-scale public programs such as USGS 3DEP~\cite{usgs_3DEP}. Because this terrain model is invalid on buildings, AeroMap3D conditions correspondences with OSM footprints~\cite{OpenStreetMap} before DEM lifting, as summarized in Table~\ref{tab:method-comparison}.

\subsection{UAV Visual Localization Datasets}

Image retrieval-based datasets such as ALTO \cite{cisneros2022alto}, UAV-VisLoc \cite{xu2024uav}, and AerialVL \cite{he2024aerialvl} focus on drone--satellite place recognition and expose strong appearance, altitude, and orientation shifts. Metric localization benchmarks provide stronger geometry: UAVD4L uses a detailed 3D reference~\cite{wu2024uavd4l}, while OrthoLoC use orthophoto--DSM references~\cite{dhaouadi2025ortholoc}. UAV-Terra3D complements them with continuous real UAV trajectories and co-registered satellite imagery, bare-earth DEM, and OSM layers, enabling evaluation of terrain-lifted 6-DoF localization with explicit semantic validity checks (Table~\ref{tab:dataset-comparison}).

\section{Problem Formulation}
\label{sec:problem-formulation}

We formulate map-anchored localization as metric pose recovery from a monocular UAV image and a public geospatial map
\begin{equation}
    \mathcal{M}=\left(I_{\mathrm{sat}}, h_{\mathrm{DEM}}, \mathcal{B}, \phi\right),
\end{equation}
where $I_{\mathrm{sat}}$ is the satellite image, $h_{\mathrm{DEM}}:\mathbb{R}^2\rightarrow\mathbb{R}$ is the bare-earth elevation field, $\mathcal{B}$ is the OSM-derived building mask in the map image plane, and $\phi$ maps a map pixel to local East-North-Up (ENU) ground coordinates.

At time $t$, the UAV body pose is $\mathbf{T}_{\mathbf{W}\mathbf{B}}^t\in\mathrm{SE}(3)$ in the world frame $\{\mathbf{W}\}$. With known gimbal extrinsics $\mathbf{T}_{\mathbf{B}\mathbf{C}}^t$, the camera pose is
\begin{equation}
    \mathbf{T}_{\mathbf{W}\mathbf{C}}^t = \mathbf{T}_{\mathbf{W}\mathbf{B}}^t \cdot \mathbf{T}_{\mathbf{B}\mathbf{C}}^t.
    \label{eq:TWC_relation}
\end{equation}
Equivalently, $\mathbf{T}_{\mathbf{C}\mathbf{W}}^t=(\mathbf{T}_{\mathbf{W}\mathbf{C}}^t)^{-1}$ maps world points into the camera frame; we write its $3\times4$ projection block as $[\mathbf{R}^t\mid\mathbf{t}^t]$.

\vspace{0.5em}
\noindent\textbf{Observation Model.}
For a lifted map point $\mathbf{X}_i\in\mathbb{R}^3$ and calibrated intrinsics $\mathbf{K}$, the ideal image measurement $\mathbf{x}_i\in\mathbb{R}^2$ satisfies
\begin{equation}
    \lambda_i
    \begin{bmatrix}\mathbf{x}_i \\ 1\end{bmatrix}
    =
    \mathbf{K}
    \begin{bmatrix}\mathbf{R}^t\!\mid\!\mathbf{t}^t\end{bmatrix}
    \begin{bmatrix}\mathbf{X}_i \\ 1\end{bmatrix}.
    \label{eq:pinhole_model}
\end{equation}
Thus the problem reduces to constructing reliable 2D--3D correspondences and solving for $\mathbf{T}_{\mathbf{C}\mathbf{W}}^t$.

\vspace{0.5em}
\noindent\textbf{Map Lifting and Structural Uncertainty.}
Let dense registration return 2D--2D matches
$\mathcal{C}_{2\mathrm{D}}^t=\{(\mathbf{x}_i^t,\mathbf{m}_i^t)\}_{i=1}^{N_t}$ between the UAV image and the satellite map. Each map pixel is lifted through
\begin{equation}
    \begin{aligned}
    (E_i,N_i) &= \phi(\mathbf{m}_i^t), \\
    \mathbf{X}_i &\doteq \ell(\mathbf{m}_i^t)
    =
    \begin{bmatrix}
        E_i & N_i & h_{\mathrm{DEM}}(E_i,N_i)
    \end{bmatrix}^{\top}.
    \end{aligned}
    \label{eq:dem_lifting}
\end{equation}
This lift is exact for terrain but biased for elevated structures. If the true surface height differs by $\Delta h_i$, then
\begin{equation}
    \mathbf{X}_i^{\mathrm{true}}
    =\ell(\mathbf{m}_i^t)+
    \begin{bmatrix}0&0&\Delta h_i\end{bmatrix}^{\top},
    \qquad
    \Delta h_i \neq 0 ,
    \label{eq:height_bias}
\end{equation}
and the resulting reprojection residual is no longer a random matching error but a structured geometric bias.

\vspace{0.5em}
\noindent\textbf{Robust Pose Objective.}
AeroMap3D first removes known structural regions in the map plane,
\begin{equation}
    \mathcal{C}_{\mathrm{sem}}^t =
    \left\{
    (\mathbf{x}_i^t,\mathbf{X}_i)
    \;\middle|\;
    (\mathbf{x}_i^t,\mathbf{m}_i^t)\in\mathcal{C}_{2\mathrm{D}}^t,
    \mathbf{m}_i^t\notin\mathcal{B}
    \right\}.
    \label{eq:semantic_filter}
\end{equation}
Residual outliers from missing OSM annotations, dynamic objects, or feature mismatches are handled by geometric consensus. With
\begin{equation}
    \mathbf{r}_i(\mathbf{T}) =
    \mathbf{x}_i^t -
    \Pi\!\left(\mathbf{K}\begin{bmatrix}\mathbf{R}\!\mid\!\mathbf{t}\end{bmatrix}
    \begin{bmatrix}\mathbf{X}_i \\ 1\end{bmatrix}\right),
\end{equation}
where $\Pi([u,v,w]^\top)=[u/w,v/w]^\top$ and $\mathbf{T}\in\mathrm{SE}(3)$ is represented by $(\mathbf{R},\mathbf{t})$. RANSAC--PnP selects an inlier index set $\mathcal{I}_t^\star$ from $\mathcal{C}_{\mathrm{sem}}^t$ and refines the camera pose by
\begin{equation}
    \hat{\mathbf{T}}_{\mathbf{C}\mathbf{W}}^t
    =
    \underset{\mathbf{T}\in\mathrm{SE}(3)}{\arg\min}
    \sum_{i\in\mathcal{I}_t^\star}
    \left\|\mathbf{r}_i(\mathbf{T})\right\|_2^2.
    \label{eq:robust_pnp}
\end{equation}
The corresponding UAV body pose is
\begin{equation}
    \hat{\mathbf{T}}_{\mathbf{W}\mathbf{B}}^t
    =
    \left(\hat{\mathbf{T}}_{\mathbf{C}\mathbf{W}}^t\right)^{-1}
    \left(\mathbf{T}_{\mathbf{B}\mathbf{C}}^t\right)^{-1},
    \label{eq:body_pose_update}
\end{equation}
which gives the metric map-based pose measurement used by the trajectory estimator.

\begin{figure}[t]
    \centering
    \includegraphics[width=\linewidth]{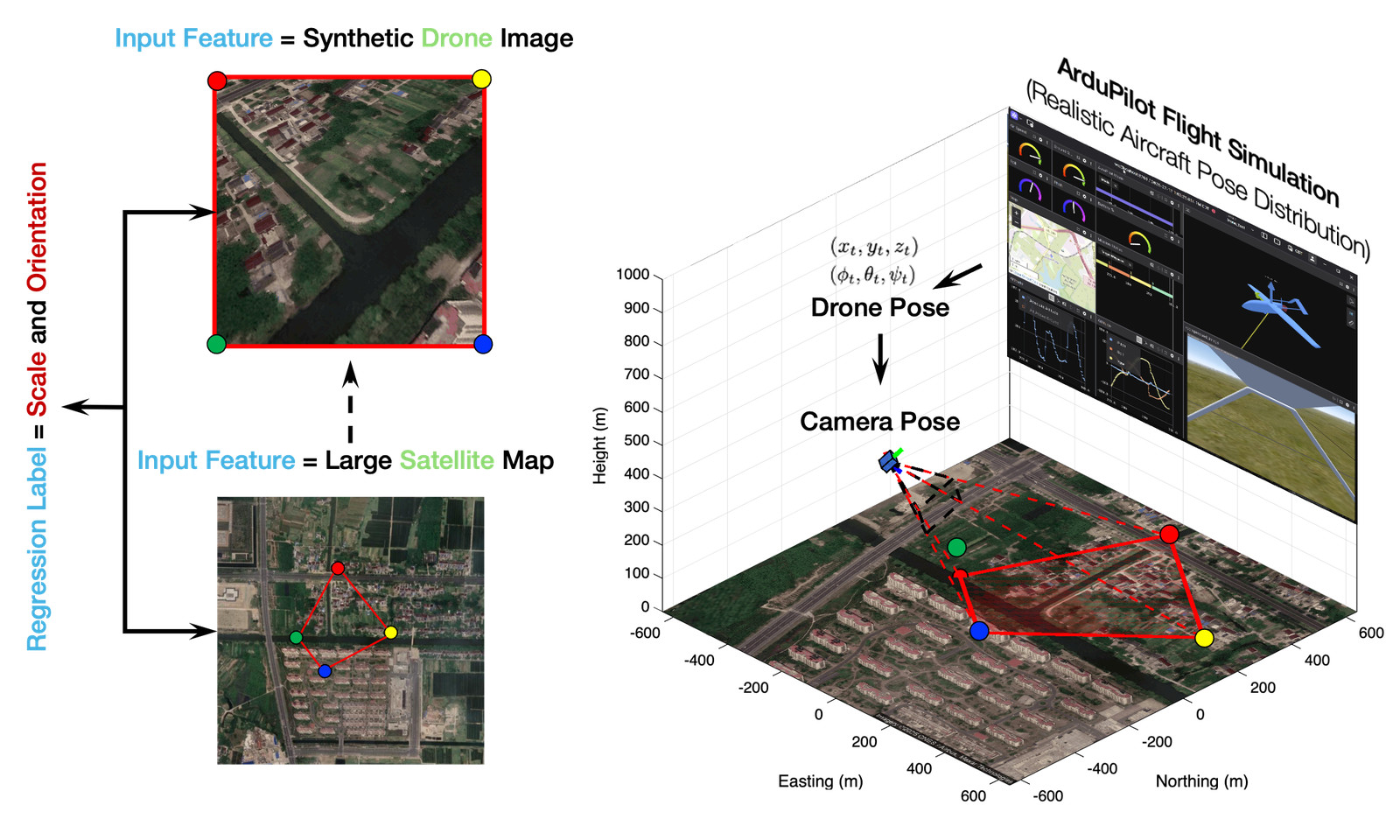}
    \caption{\textbf{Synthetic training data for the scale--yaw adapter.}
    Training UAV--map pairs are generated by applying known transformations to satellite images, providing direct supervision without real UAV images or manual labeling.}
    \label{fig:synthetic-data-generation}
    \vspace{-1.5em}
\end{figure}

\begin{figure*}[t]
    \centering
    \includegraphics[width=\linewidth]{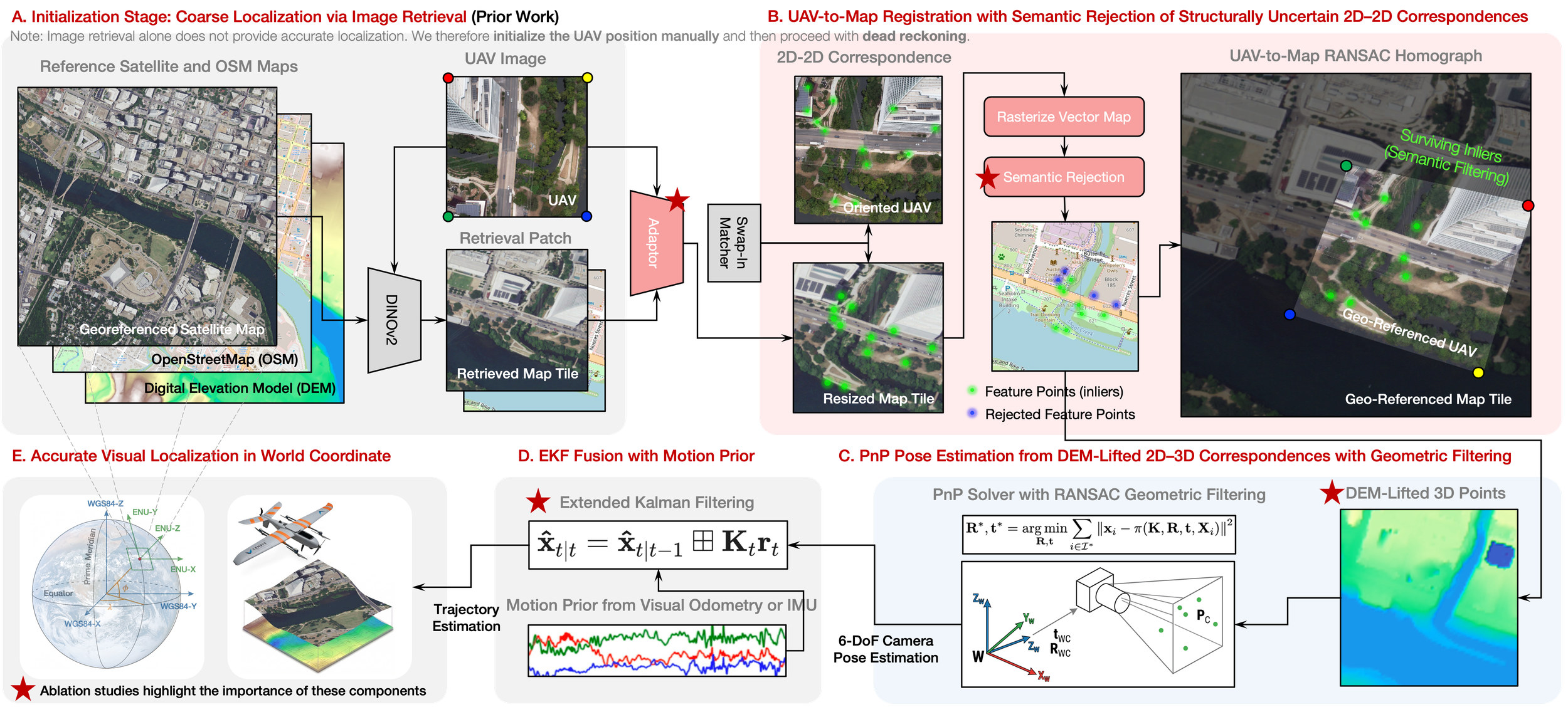}
    \caption{\textbf{AeroMap3D localization pipeline.} Starting from a coarse map initialization, obtained by retrieval in prior work or by an operator in our experiments, the system propagates with a relative-motion prior. A scale/yaw adapter enables dense UAV--map matching; OSM masking and DEM lifting form semantically reliable 2D--3D correspondences; RANSAC--PnP estimates the global camera pose; and a timestamp-aligned EKF fuses map updates with relative odometry.}
    \label{fig:pipeline}
    \vspace{-1.5em}
\end{figure*}

\section{Methodology}
\label{sec:methodology}

\subsection{Cross-View Registration via Lightweight Adapter}
\label{subsec:adapter_registration}

\noindent\textbf{Scale--Yaw Normalization.}
The registration frontend converts a difficult UAV--satellite pair into a geometry range where a pretrained matcher can operate reliably. Instead of replacing the matcher, we estimate only the dominant cross-view nuisance parameters: the scale ratio induced by altitude and field of view, and the yaw offset induced by UAV heading. This yaw is a relative image-to-map rotation used for registration, not a replacement for the vehicle's full attitude estimate. We implement the adapter as a lightweight Siamese CNN: the UAV image and satellite tile are passed through a shared MobileNetV3-Small encoder~\cite{howard2019searching}, global-average-pooled into descriptors $\mathbf{g}_t$ and $\mathbf{g}_{\mathrm{sat}}$, fused as $[\mathbf{g}_t,\mathbf{g}_{\mathrm{sat}},|\mathbf{g}_t-\mathbf{g}_{\mathrm{sat}}|,\mathbf{g}_t\odot\mathbf{g}_{\mathrm{sat}}]$, and regressed by a two-layer MLP. For UAV image $I_t$ and a north-up satellite tile $I_{\mathrm{sat}}$, the adapter predicts
\begin{equation}
    (\hat{s}_t,\hat{\theta}_t)
    =
    f_{\boldsymbol{\theta}}(I_t,I_{\mathrm{sat}}),
\end{equation}
where the MLP outputs log-scale and a normalized yaw vector internally, which is converted to $(\hat{s}_t,\hat{\theta}_t)$ for rectification. It is trained with
\begin{equation}
    \mathcal{L}_{\text{align}}
    =
    \left|\log\hat{s}_t-\log s_t^\star\right|
    +
    \beta\left(1-\cos(\hat{\theta}_t-\theta_t^\star)\right),
    \label{eq:alignment_loss}
\end{equation}
where $(s_t^\star,\theta_t^\star)$ are synthetic supervision targets and $\beta$ balances scale and yaw. Training pairs are generated without manual correspondence labels or UAV-Terra3D images: we sample plausible flight states, project their camera footprints onto satellite imagery, and render UAV-like views with known scale and yaw (Fig.~\ref{fig:synthetic-data-generation}).

This supervision normalizes scale and yaw but does not simulate seasonal appearance transfer or full projective distortion from large roll or pitch; these remain limitations of the frozen matcher and map reference.

\vspace{0.5em}
\noindent\textbf{Dense Matching After Rectification.}
The predicted $(\hat{s}_t,\hat{\theta}_t)$ rectifies the map tile before matching. We then run RoMav2~\cite{edstedt2025roma} as a swap-in dense matcher and map the matched satellite pixels back to the original map plane:
\begin{equation}
    \mathcal{C}_{2\mathrm{D}}^t
    =
    \left\{
    \bigl(\mathbf{x}_i^t,\mathbf{m}_i^t\bigr)
    \right\}_{i=1}^{N_t}.
    \label{eq:method_2d_matches}
\end{equation}
Here $\mathbf{x}_i^t$ is a UAV image pixel and $\mathbf{m}_i^t$ is its satellite-map pixel. This keeps the dense matcher modular while producing the 2D--2D correspondences that are subsequently lifted onto the DEM and filtered with semantic map priors.

\subsection{Semantic Map Conditioning for DEM-PnP}
\label{subsec:semantic-conditioning}

\noindent\textbf{Semantic Filtering Before 3D Lifting.}
The core contribution of this stage is semantic conditioning before pose estimation. A bare-earth DEM provides scalable terrain geometry, but it assigns terrain height to every map pixel; matches on roofs, facades, or other above-ground structures therefore inherit the height bias in Eq.~\eqref{eq:height_bias}. Because these errors are coherent rather than random, RANSAC alone can accept an incorrect dominant surface. AeroMap3D instead removes known non-ground matches before DEM lifting. OSM building footprints are rasterized into the binary mask $\mathcal{B}$, and candidate matches on these semantically unreliable regions are rejected in the map plane:
\begin{equation}
    \mathcal{C}_{\mathrm{gnd}}^t
    =
    \left\{
    (\mathbf{x}_i^t,\mathbf{m}_i^t)
    \in \mathcal{C}_{2\mathrm{D}}^t
    \;\middle|\;
    \mathbf{m}_i^t\notin\mathcal{B}
    \right\}.
\end{equation}
Only the surviving map pixels are lifted with $\ell(\cdot)$ from Eq.~\eqref{eq:dem_lifting}, yielding terrain-supported 2D--3D correspondences and preventing known structural regions from becoming confidently wrong 3D points.

\vspace{0.5em}
\noindent\textbf{Geometric Consensus for Residual Outliers.}
Semantic filtering removes the dominant map-induced bias but cannot cover missing OSM annotations, off-nadir building lean, transient objects, or dense-matcher failures. These residual errors are handled by standard OpenCV RANSAC--PnP~\cite{opencv_solvepnp_forum}, corresponding to Eq.~\eqref{eq:robust_pnp}; the novelty is the semantic conditioning that supplies PnP with terrain-consistent correspondences. The resulting metric pose is the global visual measurement passed to the delayed-update EKF in Sec.~\ref{subsec:ekf_fusion}.

This residual-consensus stage still assumes that terrain-consistent correspondences remain dominant; a large unmapped structure can form a coherent but incorrect consensus and remains a map-validity failure mode.

\subsection{Continuous Trajectory Estimation via Map-Anchored EKF}
\label{subsec:ekf_fusion}

\noindent\textbf{Pose-Error State and Motion Prior.}
The nominal EKF state is the map-frame body pose $\hat{\mathbf{T}}_k\in\mathrm{SE}(3)$ with right-multiplicative local error $\delta\boldsymbol{\xi}_k\in\mathbb{R}^6$, i.e., $\mathbf{T}_k=\hat{\mathbf{T}}_k\operatorname{Exp}(\delta\boldsymbol{\xi}_k^\wedge)$, and tangent-space covariance $\mathbf{P}_k$. Between map updates, it propagates with the relative pose increment $\tilde{\Delta\mathbf{T}}_k$ from Sec.~\ref{sec:dataset}:
\begin{equation}
    \begin{aligned}
    \hat{\mathbf{T}}_{k|k-1}
    &= \hat{\mathbf{T}}_{k-1|k-1}\tilde{\Delta\mathbf{T}}_k,\\
    \mathbf{P}_{k|k-1}
    &= \mathbf{F}_k\mathbf{P}_{k-1|k-1}\mathbf{F}_k^\top+\mathbf{Q}_k .
    \end{aligned}
    \label{eq:ekf_prediction}
\end{equation}
Here $\mathbf{F}_k$ and $\mathbf{Q}_k$ are the pose-composition Jacobian and motion-increment covariance. PnP provides $\mathbf{Z}_k\in\mathrm{SE}(3)$ from Eq.~\eqref{eq:body_pose_update}, with tangent-space covariance $\mathbf{R}_k$.

Visual poses are tagged at image acquisition. The reported offline experiments fuse them in timestamp order and do not evaluate wall-clock delay. For online use, a pose captured at $k$ but returned at $j\ge k$ updates the buffered state at $k$; stored motion increments then re-propagate the estimate to $j$.

\vspace{0.5em}
\noindent\textbf{Mahalanobis Gating.}
For the delayed state, we compute the innovation and its consistency score
\begin{equation}
    \begin{aligned}
    \mathbf{r}_k
    &= \operatorname{Log}\!\left(
       \hat{\mathbf{T}}_{k|k-1}^{-1}\mathbf{Z}_k
       \right)^\vee,\\
    \mathbf{S}_k
    &= \mathbf{H}_k\mathbf{P}_{k|k-1}\mathbf{H}_k^\top+\mathbf{R}_k,\\
    \gamma_k
    &= \mathbf{r}_k^\top\mathbf{S}_k^{-1}\mathbf{r}_k .
    \end{aligned}
    \label{eq:ekf_gate}
\end{equation}
where $(\cdot)^\vee$ gives 6-vector tangent coordinates and $\mathbf{H}_k=\mathbf{I}_6$. The visual pose is accepted only if
\begin{equation}
    \gamma_k < \tau_{\chi^2},
    \qquad
    \tau_{\chi^2}=\chi^2_{d,1-\eta},
    \label{eq:chi2_gate}
\end{equation}
where $d$ is the measurement dimension and $\eta$ is the gate probability. Rejected measurements are discarded and the trajectory coasts on the relative-motion prior. Accepted measurements use the standard EKF update
\begin{equation}
    \begin{aligned}
    \mathbf{K}_k
    &= \mathbf{P}_{k|k-1}\mathbf{H}_k^\top\mathbf{S}_k^{-1},\\
    \delta\hat{\boldsymbol{\xi}}_k
    &= \mathbf{K}_k\mathbf{r}_k,\\
    \hat{\mathbf{T}}_{k|k}
    &= \hat{\mathbf{T}}_{k|k-1}
       \operatorname{Exp}(\delta\hat{\boldsymbol{\xi}}_k^\wedge),\\
    \mathbf{P}_{k|k}
    &= (\mathbf{I}-\mathbf{K}_k\mathbf{H}_k)\mathbf{P}_{k|k-1}
       (\mathbf{I}-\mathbf{K}_k\mathbf{H}_k)^\top
       +\mathbf{K}_k\mathbf{R}_k\mathbf{K}_k^\top .
    \end{aligned}
    \label{eq:ekf_update}
\end{equation}
The corrected state is then re-propagated; $\operatorname{Prop}_\ell$ applies the stored increment $\tilde{\Delta\mathbf{T}}_\ell$:
\begin{equation}
    \begin{aligned}
    (\hat{\mathbf{T}}_{\ell|\ell-1},\mathbf{P}_{\ell|\ell-1})
    &=
    \operatorname{Prop}_\ell\!\left(
    \hat{\mathbf{T}}_{\ell-1|\ell-1},
    \mathbf{P}_{\ell-1|\ell-1}
    \right),\\
    \ell&=k+1,\ldots,j .
    \end{aligned}
    \label{eq:delayed_repropagation}
\end{equation}

\begin{figure*}[t]
    \centering
    \includegraphics[width=\linewidth]{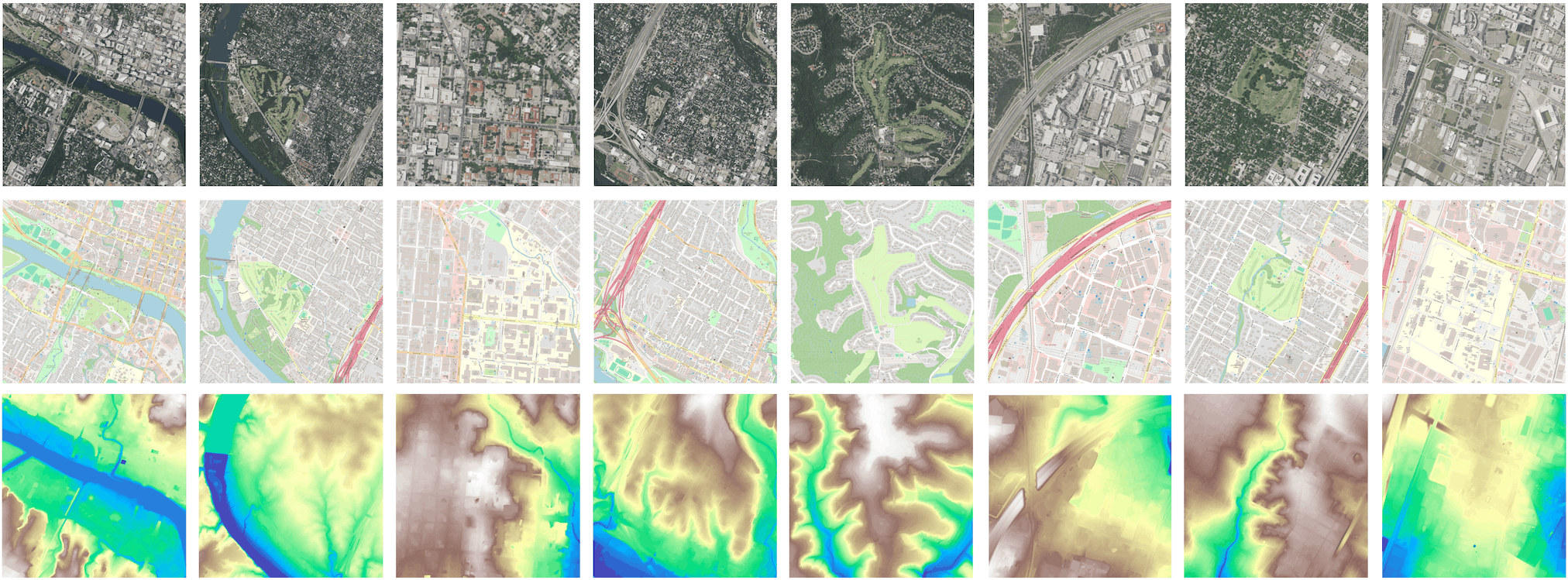}
    \caption{\textbf{UAV-Terra3D map priors.}
    Eight Austin sites combine WGS84-aligned NAIP imagery~\cite{usda_naip} for registration (top; 1~m/pixel), OpenStreetMap~\cite{OpenStreetMap} for semantic filtering (middle), and USGS DEMs~\cite{usgs_3DEP} for 2D--3D lifting (bottom; 1~m/pixel).}
    \label{fig:dataset}
    \vspace{-1.5em}
\end{figure*}

\section{Dataset}
\label{sec:dataset}

\noindent\textbf{UAV-Terra3D Benchmark.}
UAV-Terra3D is a benchmark for map-anchored UAV visual localization. It contains 30-Hz, mostly nadir-view UAV video, calibrated camera intrinsics, GNSS position reference trajectories, and aligned visual, geometric, and semantic map layers. We collected 20 trajectories at eight sites in Austin, Texas, using a DJI Air 2S equipped with a calibrated gimballed camera. The dataset covers 55~km and 134~minutes of flight over 22.4~km$^2$, with varied altitudes, headings, and environments. Most frames are nadir or near-nadir; the relatively few high-pitch frames do not support a statistically meaningful pitch-stratified registration analysis. The collection device does not provide independently validated full 6-DoF ground truth or gimbal-encoder calibration suitable for evaluating roll, pitch, and yaw error.

\noindent\textbf{3D Terrain Models.}
Each site combines NAIP imagery~\cite{usda_naip}, USGS DEM elevation~\cite{usgs_3DEP}, and OSM semantics~\cite{OpenStreetMap} in a WGS84-referenced local ENU frame (Fig.~\ref{fig:dataset}). The 1-m 3DEP tiles are horizontally NAD83 and vertically NAVD88; we transform them to WGS84 and convert orthometric elevation $H$ to ellipsoidal height $h=H+N$ using the metadata geoid separation $N$, matching the GNSS altitude convention. The imagery supports registration, the DEM supplies 3D terrain, and OSM identifies regions where bare-earth elevation is unreliable.

\noindent\textbf{Motion-Prior Generation.}
UAV-Terra3D does not include a raw IMU stream. To compare trajectory methods under the same realistic, imperfect motion input, we derive a reproducible stochastic odometry prior from consecutive GNSS reference poses $\mathbf{T}_{t-1}^\star,\mathbf{T}_t^\star\in\mathrm{SE}(3)$. We perturb each relative motion by zero-mean noise whose scale increases with translation and rotation, thereby modeling the accumulation of odometry uncertainty without using GNSS as an absolute EKF update:
\begin{equation}
    \begin{aligned}
    \Delta \mathbf{T}_t^\star =
    \left(\mathbf{T}_{t-1}^\star\right)^{-1}\mathbf{T}_t^\star,\\
    \tilde{\Delta \mathbf{T}}_t =
    \Delta \mathbf{T}_t^\star \operatorname{Exp}(\boldsymbol{\xi}_t),\\
    \boldsymbol{\xi}_t \sim \mathcal{N}(\mathbf{0}, \boldsymbol{\Sigma}_t).
    \end{aligned}
    \label{eq:stochastic_odometry}
\end{equation}
Following the probabilistic odometry model~\cite{thrun2005probabilistic}, the translation and rotation noise scales are
\begin{equation}
    \begin{aligned}
    \boldsymbol{\Sigma}_t
    &= \operatorname{diag}\!\left(
    \sigma_{p,t}^2\mathbf{1}_3,
    \sigma_{R,t}^2\mathbf{1}_3
    \right),\\
    \sigma_{p,t}
    &= \alpha_3\|\Delta \mathbf{p}_t^\star\|
    + \alpha_4\|\operatorname{Log}(\Delta \mathbf{R}_t^\star)\|
    + \beta_p,\\
    \sigma_{R,t}
    &= \alpha_1\|\operatorname{Log}(\Delta \mathbf{R}_t^\star)\|
    + \alpha_2\|\Delta \mathbf{p}_t^\star\|
    + \beta_R .
    \end{aligned}
    \label{eq:odometry_noise}
\end{equation}
Here $\Delta \mathbf{p}_t^\star$ and $\Delta \mathbf{R}_t^\star$ are the translational and rotational components of $\Delta \mathbf{T}_t^\star$; parameter values are given in Sec.~\ref{sec:experiments}. GNSS is used only to generate this fixed prior and to score accuracy, never as an absolute localization update. In addition to this reference-derived relative-motion prior, we generate a visual-only prior with frame-to-frame ORB visual odometry~\cite{nister2004visual,rublee2011orb}.

\section{Experiments}
\label{sec:experiments}
We conduct real-world experiments to answer four questions: (Q1) Can an adapter trained only on synthetic UAV-VisLoc data collected in China transfer to real UAV imagery from the geographically distinct UAV-Terra3D dataset in the United States, without target-dataset retraining? (Q2) How much does the adapter improve UAV-to-map registration when paired with a frozen image matcher? (Q3) Does geometric--semantic correspondence filtering improve single-frame 6-DoF camera pose estimation over geometric-only PnP and image-retrieval baselines? (Q4) Can fusing map-based visual updates with motion priors reduce localization error and bound long-horizon trajectory drift?

\subsection{Experimental Setup}
\label{subsec:experimental-setup}
\noindent\textbf{Protocol and Metrics.}
The adapter is trained only on synthetic UAV--map pairs rendered from UAV-VisLoc satellite imagery~\cite{xu2024uav} (Fig.~\ref{fig:synthetic-data-generation}) and evaluated by cross-dataset transfer to UAV-Terra3D; no target data are used for training or tuning. All evaluations are offline and start from a user-provided initial $(x,y)$ position. We evaluate every UAV-Terra3D trajectory and frame without site-, trajectory-, or frame-level cherry-picking. Visual poses retain their acquisition timestamps and are fused in timestamp order; Sec.~\ref{subsec:ekf_fusion} describes delayed-state handling for online processing.

For image registration, we report pseudo success rate (i.e., the fraction of frames with sufficient inliers and a valid fundamental matrix or homography after geometric verification) and runtime (ms/frame). For trajectories, we report the success rate within 50~m and absolute 3D localization error.

\begin{figure}[t]
    \centering
    \includegraphics[width=\linewidth]{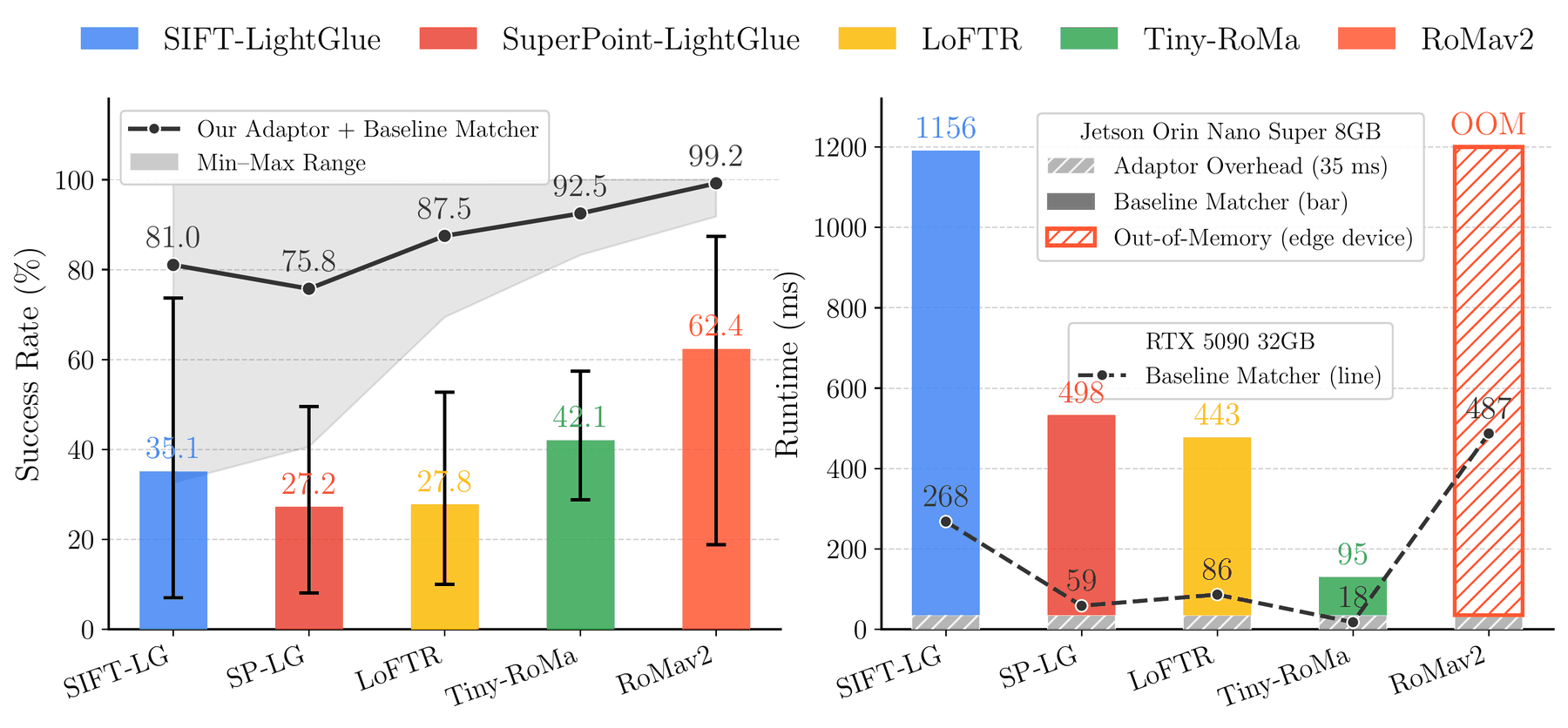}
    \caption{\textbf{Cross-domain UAV-to-map registration and runtime.}
    \textbf{Left:} Direct-matcher success (bars; I-shaped error bars indicate the minimum--maximum range across the dataset) versus adapter-enhanced success (line). Our adapter raises RoMav2 from $62.4\%$ to $99.2\%$ and Tiny-RoMa from $42.1\%$ to $92.5\%$. \textbf{Right:} Jetson runtimes (bars, including $35$\,ms adapter overhead) and RTX~5090 runtimes (dashed line); RoMav2 exceeds the Jetson memory budget.}
    \label{fig:registration_Accuracy}
    \vspace{-1.5em}
\end{figure}

\noindent\textbf{Implementation Details.}
The scale--yaw adapter is a Siamese MobileNetV3-Small (Sec.~\ref{subsec:adapter_registration}) trained on ten rendered UAV-like views from each of 3,394 UAV-VisLoc scenes. It takes $224 \times 224$ UAV and satellite inputs. Our default frozen matcher, RoMav2~\cite{edstedt2025roma}, uses $518 \times 518$ inputs, 1,024 keypoints, and ImageNet normalization. We estimate poses with OpenCV PnP and reprojection-error refinement, optimize world points in local ENU coordinates, and convert the result to WGS84. For the reference-derived motion prior in Eq.~\eqref{eq:odometry_noise}, we use $(\alpha_1,\alpha_2,\alpha_3,\alpha_4)=(0.05,0.01\,\mathrm{rad\,m^{-1}},0.05,0.10\,\mathrm{m\,rad^{-1}})$, $\beta_p=0.05$~m, and $\beta_R=0.5^\circ$; the EKF uses a 95\% $\chi^2$ gate for 6-DoF updates ($\tau_{\chi^2}=12.59$).

\noindent\textbf{Baselines.}
\textit{(1) UAV-to-map image registration.} We compare SIFT+LightGlue~\cite{lindenberger2023lightglue}, SuperPoint+LightGlue~\cite{detone2018superpoint}, LoFTR~\cite{sun2021loftr}, and RoMa matchers~\cite{edstedt2024roma,edstedt2025roma}. Each is evaluated directly and after adapter-based scale--yaw normalization, with matcher weights fixed.
\textit{(2) Single-frame camera pose estimation.} We compare DINOv2-based retrieval~\cite{yang2025dinov2} and SelaVPR++~\cite{lu2025selavpr++}, using $150 \times 150$~m satellite tiles sampled at a 10~m stride as the candidate database. We also include planar 2D-map homography, sparse DEM-lifted PnP from four UAV-footprint corners, and a dense-PnP ablation without semantic rejection.
\textit{(3) Multi-frame trajectory estimation.} We compare Ref-Odom in Eq.~(\ref{eq:stochastic_odometry}), ORB visual odometry~\cite{nister2004visual,rublee2011orb}, ORB-SLAM2~\cite{mur2017orb}, and GeoVINS-style retrieval~\cite{li2025geovins}. The first three have no satellite-map anchoring. Our GeoVINS-style baseline uses DINOv2~\cite{oquab2023dinov2} for coarse retrieval updates and shares the same initialization, noisy motion prior, and EKF interface as AeroMap3D.

\begin{table}[t]
\centering
\begin{threeparttable}
\scriptsize
\sffamily
\renewcommand{\arraystretch}{1.1}
\begin{tabularx}{\columnwidth}{@{} X S[table-format=3.2] S[table-format=2.2] S[table-format=2.2] @{}}
\toprule
\text{Method} & {\text{Success (\%)}} & {\text{Mean 3D (m)}} & {\text{95th 3D (m)}} \\
\midrule
\multicolumn{4}{@{}l}{\textit{Single-Frame Visual Localization}} \\
\midrule
DINOv2-based \cite{yang2025dinov2}              & 16.47  & 29.48 & 47.58 \\
SelaVPR++ \cite{lu2025selavpr++}          & 19.35  & 29.30 & 47.65 \\
2D Map + Homography\tnote{3}                   & 62.20  & 30.80 & 44.70 \\
Sparse DEM-PnP (Footprint)\tnote{4}             & 44.70  & 23.90 & 45.40 \\
Dense DEM-PnP (RANSAC)\tnote{5}               & 88.24  & 16.17 & 32.27 \\
{\bfseries Ours} (RANSAC--PnP + OSM)    & {\bfseries 95.69}  & {\bfseries 14.11} & {\bfseries 28.20} \\
\midrule
\multicolumn{4}{@{}l}{\textit{Continuous Trajectory Estimation}} \\
\midrule
Ref-Odom \tnote{6} \cite{thrun2005probabilistic}   & 14.90  & 24.20 & 45.14 \\
ORB-Visual Odometry \tnote{7}\cite{nister2004visual,rublee2011orb}   & 45.61  & 28.24 & 47.90 \\
ORB-SLAM2 \tnote{8} \cite{mur2017orb}                   & 41.44  & 14.71 & 37.31 \\
GeoVINS-style retrieval \tnote{9} \cite{li2025geovins}                    & {\bfseries 100.00} & 17.17 & 34.78 \\
\bfseries Ours (PnP+EKF)                      & {\bfseries 100.00} & {\bfseries 5.88} & {\bfseries 11.25} \\
\bottomrule
\end{tabularx}
\begin{tablenotes}
\item[3] Planar 2D-map homography; no elevation data.
\item[4] Sparse PnP from four DEM-lifted UAV-footprint corners.
\item[5] Dense DEM-lifted RANSAC--PnP with only the OSM filter disabled.
\item[6] Reference-derived stochastic propagation without visual or map updates.
\item[7] Monocular planar visual odometry with an ORB frontend.
\item[8] Monocular ORB-SLAM2 with Ref-Odom propagation and no map anchoring.
\item[9] DINOv2-based GeoVINS-style classify-then-retrieve updates fused with the same Ref-Odom prior and EKF interface as AeroMap3D.
\item[] \textit{Metrics:} success denotes 3D error below $50$~m; mean and 95th-percentile errors are computed over successful frames.
\end{tablenotes}
\caption{\textbf{Single-frame and continuous localization performance.} Although AeroMap3D estimates full 6-DoF poses, UAV-Terra3D provides ground truth only for the 3-DoF position component. Therefore, the evaluation is restricted to translational localization accuracy, while rotational accuracy is not reported.}
\label{tab:localization-performance}
\end{threeparttable}
\vspace{-1.5em}
\end{table}

\begin{figure}[t]
    \centering
    \includegraphics[width=\linewidth]{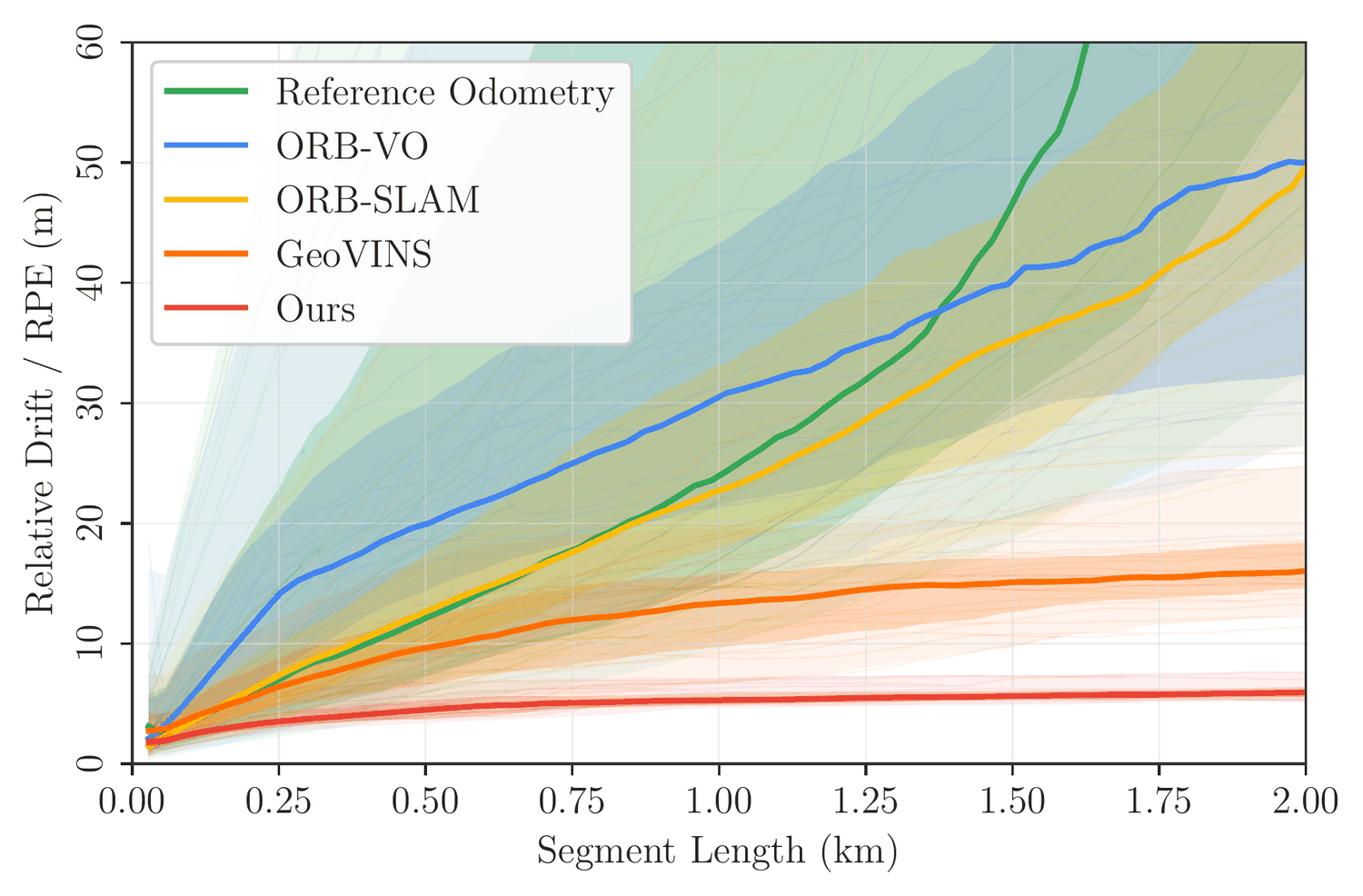}
    \caption{\textbf{Long-horizon relative drift.} RPE versus segment length; lines show medians, shaded regions show the 25--75\% and 5--95\% ranges, and faint traces show individual trajectories. AeroMap3D remains below $6$~m RPE at 2~km, while reference odometry, ORB-VO, and ORB-SLAM2 accumulate substantial drift.}
    \label{fig:error_band}
    \vspace{-1.5em}
\end{figure}

\begin{figure}[t]
    \centering
    \includegraphics[width=\linewidth]{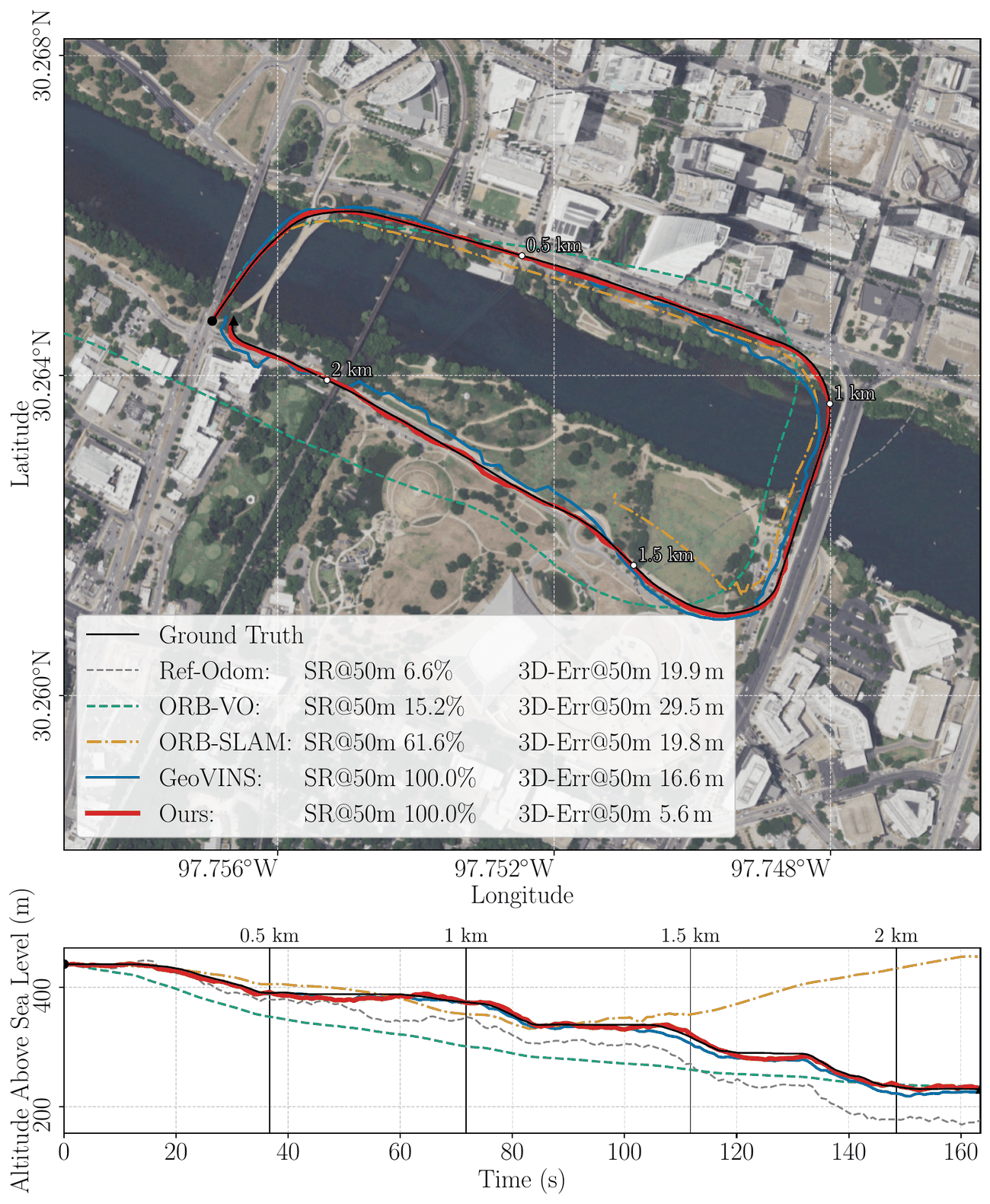}
    \caption{\textbf{Representative 2-km UAV-Terra3D trajectory.} \textbf{Top:} map-space trajectories. \textbf{Bottom:} altitude profiles. AeroMap3D closely follows the ground truth while map-unanchored odometry and SLAM drift; GeoVINS bounds drift but remains less accurate than metric PnP updates.}
    \label{fig:trajectory_comparison}
    \vspace{-1.5em}
\end{figure}

\subsection{Q1 \& Q2: Cross-Dataset Registration}
\label{subsec:registration-results}
Fig.~\ref{fig:registration_Accuracy} evaluates cross-dataset transfer from synthetic UAV-VisLoc training pairs to real UAV-Terra3D imagery without target-dataset training. Despite the scale- and orientation-robust features of SIFT and RoMa, both fall short under the large cross-view appearance, footprint-scale, and yaw mismatch: SIFT+LightGlue reaches $35.1\%$ geometric-verification pass rate and RoMav2 reaches $62.4\%$. The adapter normalizes scale and yaw before correspondence extraction and improves every evaluated matcher. In particular, Adapter+RoMav2 reaches $99.2\%$, a $36.8$-point gain, providing the geometrically consistent correspondences required for DEM-lifted PnP.

On the RTX~5090, Tiny-RoMa is the fastest configuration (18~ms), whereas RoMav2 delivers the highest registration accuracy but requires 487~ms. For edge deployment, we recommend Adapter+Tiny-RoMa: it reaches $92.5\%$ success on a Jetson Orin Nano in 130~ms (35~ms adapter plus 95~ms matcher), while RoMav2 exceeds the device's memory budget. The delayed-update EKF accommodates these asynchronous visual updates.

\subsection{Q3: Single-Frame Camera Pose Estimation}
\label{subsec:pose-results}
The upper half of Table~\ref{tab:localization-performance} shows that retrieval provides only coarse localization. SelaVPR++, whose training is tailored to visual place recognition, improves on the DINOv2-based method but still reaches only $19.35\%$ success and 29.30~m mean error. The planar baseline estimates a RANSAC homography between the UAV image and the geo-referenced map, then decomposes it using camera intrinsics and the map scale to recover pose relative to the ground plane. It improves success to $62.20\%$ but retains 30.80~m error, demonstrating that the planar assumption is insufficient. Sparse DEM-lifted PnP reduces the mean error to 23.90~m but succeeds on only $44.70\%$ of frames because footprint corners frequently fall on elevated or unmapped structures.

To isolate semantic conditioning, we disable only the OSM pre-filter while holding the matcher, DEM lifting, PnP solver, RANSAC thresholds, and evaluation set fixed. Dense DEM-PnP reaches $88.24\%$ success and 16.17~m mean error; adding OSM raises success to $95.69\%$ and reduces the mean and 95th-percentile errors to 14.11~m and 28.20~m. This controlled comparison measures the incremental benefit of semantic conditioning before standard geometric consensus.

\subsection{Q4: Continuous Trajectory Estimation}
\label{subsec:trajectory-results}
The lower half of Table~\ref{tab:localization-performance} and Figs.~\ref{fig:error_band}--\ref{fig:trajectory_comparison} show that motion propagation alone or monocular visual odometry cannot prevent long-range drift: Ref-Odom reaches only $14.90\%$ success, and ORB-VO and ORB-SLAM2 remain unanchored to the map. GeoVINS-style retrieval achieves $100\%$ success with 17.17~m mean error, whereas AeroMap3D preserves $100\%$ success with 5.88~m mean and 11.25~m 95th-percentile error. This $65.8\%$ lower mean error, together with bounded RPE in Fig.~\ref{fig:error_band} and the representative trajectory in Fig.~\ref{fig:trajectory_comparison}, shows the benefit of metric map updates.

\section{Conclusion}
\label{sec:conclusion}

We presented AeroMap3D for map-anchored UAV localization from satellite imagery, bare-earth DEMs, and OSM building footprints. By normalizing scale--yaw mismatch and filtering semantically invalid DEM-lifted correspondences before standard RANSAC--PnP, AeroMap3D raises RoMav2's geometric-verification pass rate from $62.4\%$ to $99.2\%$ and single-frame localization success from $88.24\%$ to $95.69\%$. With a reference-derived stochastic motion prior, the trajectory estimator achieves 5.88~m mean translation error over 55~km.

\noindent\textbf{Limitations and Future Work.} The evaluation is limited to eight Austin-area sites, coarse initialization, translation-only ground truth, and a GNSS-reference-derived motion prior. OSM gaps, dominant unmapped structures, strongly oblique views, and seasonal or temporal map changes remain failure modes. Future work will evaluate measured onboard odometry, multi-region flights, attitude and gimbal-extrinsic accuracy, pitch-stratified registration, global initialization, map-validity modeling, and uncertainty-aware visual updates.

\noindent\textbf{Acknowledgment.}
This work has taken place in the Human Centered Robotics Laboratory (HCRL) at UT Austin. This work has been made possible thanks to the generous gifts from Leo Lion and the Sentis-Ben-Yakar family.

\bibliographystyle{unsrtnat}
{\scriptsize
\setlength{\bibsep}{0pt}
\renewcommand{\baselinestretch}{1.0}\selectfont
\bibliography{references.bib}
}
\end{document}